\documentclass[runningheads]{llncs}
\usepackage[T1]{fontenc}
\usepackage{hyperref}
\usepackage{booktabs}
\usepackage{multirow}
\usepackage[most]{tcolorbox}
 \usepackage{enumitem}
\usepackage{graphicx}
\usepackage{graphicx}
\usepackage{gensymb}
\usepackage{amsmath}
\usepackage{amssymb}
\usepackage{enumitem}
\usepackage{multirow}
\usepackage{longtable}
\usepackage{adjustbox}
\usepackage{array}
\usepackage{booktabs}
\usepackage{pgfplots}
\pgfplotsset{compat=1.18}
\usepackage{pgfplotstable}
\usepackage{tikz}
\begin{document}
\pagestyle{empty}
\title{Balancing Foundation Models and Trait-Specific Learning for Forehead-Crease Recognition}
\author{Geetanjali Sharma\inst{1} \and
Aryan Jain\inst {1} \and
Dev Nagaich\inst{1} \and
Aditya Nigam\inst{1} \and
Raghavendra Ramachandra \inst{2}}
%
%
\institute{Indian Institute of Technology Mandi, India \and
SAFE Center, Norwegian University of Science and Technology (NTNU), Gjøvik, Norway
\\}

%
%
%
%
\maketitle              
\begin{abstract}
Biometric recognition often degrades under partial facial visibility, where conventional facial cues are occluded. The forehead region remains consistently visible and contains distinctive crease patterns that can serve as a complementary biometric trait. We present a unified evaluation of forehead-crease recognition across explainable zero-shot reasoning, domain-adaptive learning, and trait-specific modeling. FIRE enables interpretable zero-shot verification using multimodal large language models, while pretrained face-recognition models are adapted to the forehead domain through margin-based learning. Under open-set evaluation, the best zero-shot model 
GPT-4o+FIRE achieves 18.22\% EER, while Edge-Forehead achieves 8.30\% EER and 54.01\% TMR@0.1\% FMR. The results show that although MLLMs offer interpretable zero-shot verification, trait-specific learning remains more effective for robust forehead-crease biometric recognition.

\end{abstract}
\vspace{-0.1in}
\section{Introduction}
\label{sec:intro}
\vspace{-0.1in}
Biometric verification systems~\cite{bharadwaj2022mobile,jain2004introduction} have achieved remarkable progress in recent years, largely driven by deep learning, supervised approaches. These methods leverage large-scale annotated datasets and powerful representation learning techniques~\cite{schroff2015facenet,sharma2024fh}, enabling high recognition accuracy under controlled conditions. As a result, such systems are widely deployed in applications including access control, surveillance, and identity verification. However, their performance degrades significantly in unconstrained environments, particularly when faces are partially occluded due to masks, scarves, pose variations, or low-resolution imaging. In these scenarios, a substantial portion of the facial region becomes unavailable, limiting the effectiveness of holistic face-based representations.
To address this limitation, recent studies have explored partial facial regions and soft biometric traits that remain visible under occlusion~\cite{bharadwaj2022mobile,sharma2024fh}. Among these, the forehead is particularly promising due to its consistent visibility and rich structural characteristics. 
Forehead creases provide discriminative identity cues, with prior studies demonstrating verification across images acquired in two sessions; however, their temporal stability remains insufficiently quantified~\cite{bharadwaj2022mobile,sharma2024fh}.
Furthermore, fine-grained skin texture and wrinkle patterns retain identity-discriminative information even in the absence of global facial cues.

Despite these advantages, existing forehead-crease recognition methods face several limitations. Most approaches rely on task specific architectures and do not fully exploit transferable representations learned by modern face recognition models. Although foundation models(FMs) and multimodal large language models (MLLMs) have demonstrated strong Zero-Shot(ZS) capabilities, they are not optimized for subtle biometric cues such as crease topology, spacing, and texture continuity. In addition, prior studies often compare ZS and supervised approaches under different protocols, limiting fair evaluation. Finally, explainability in forehead-based biometric recognition remains largely underexplored despite its importance for transparency and real-world deployment.
In parallel, modern face recognition models
have been shown to learn hierarchical representations that capture not only global facial structure but also mid-level and fine-grained texture patterns~\cite{schroff2015facenet}. This observation motivates an important research question: \emph{can pretrained face recognition models be effectively adapted to operate on spatially constrained regions such as the forehead?} Recent advances in margin-based metric learning, including AdaFace~\cite{kim2022adaface}, CosFace~\cite{wang2018cosface}, and related formulations, provide a promising direction by enforcing strong intra-class compactness and inter-class separability, even under subtle appearance variations. These properties make them particularly suitable for modelling fine-scale biometric traits such as forehead creases. This work systematically compares FMs with domain-adaptive and trait-specific approaches for trustworthy and interpretable biometric verification.

Motivated by these insights, we propose a unified framework for explainable and domain-adaptive forehead-crease biometric recognition that integrates three complementary paradigms within a single system. Unlike prior work that treats these directions independently, our approach combines: \textbf{(i)} explainable zero-shot reasoning using multimodal models, \textbf{(ii)} domain adaptation of pretrained face recognition networks using margin-based metric learning, and \textbf{(iii)} a trait-specific model designed to explicitly capture fine-grained crease structures. This unified formulation enables a comprehensive understanding of the trade-off between interpretability, generalisation, and discriminative performance.
We evaluate the proposed framework under open-set verification protocols on real-world and synthetic forehead datasets. To ensure a fair and rigorous comparison, we benchmark multiple paradigms, including multimodal large language models, vision foundation models, domain-adapted face recognition systems, and specialised forehead-crease architectures. In addition, we conduct detailed ablation studies to analyse the impact of structured prompting, margin-based supervision, and training strategies on overall performance.
From an application perspective, the proposed system is particularly relevant in scenarios where full-face information is unavailable or unreliable, such as masked authentication, surveillance in unconstrained environments, and low-resolution or partial-face recognition. In such cases, forehead-crease biometrics can act as a complementary modality, improving robustness without requiring additional sensors or intrusive acquisition processes.

\textbf{Contributions.} The main contributions of this work are: (i) a unified framework integrating explainable zero-shot reasoning, domain-adaptive learning, and trait-specific modeling for forehead-crease recognition; (ii) adaptation of pretrained face-recognition models to forehead biometrics through margin-based metric learning; (iii) FIRE, a structured framework for interpretable multimodal verification; (iv) a balanced open-set evaluation of zero-shot, domain-adaptive, and trait-specific approaches; and (v) an assessment of forehead-crease biometrics for partial-face recognition scenarios.
The remainder of this paper is structured as follows. Section~\ref{sec:related} reviews related work. Section~\ref{sec:method} presents the proposed methodology. Section~\ref{sec:dataset_and_eval_protocol} describes the datasets and evaluation protocols. Section~\ref{result_discussion} presents the results and discussion, followed by additional ablation studies and prompt analyses are provided in the supplementary material. 
Finally, Section~\ref{conclusion} concludes the paper.

\section{Related work}
\label{sec:related}

In this section, we review existing literature on forehead-based biometrics, face recognition and domain adaptation, and explainable verification using multimodal models. We also highlight key limitations and position our contribution within these research directions.

\subsection{Forehead-Based Biometric Recognition}

Forehead-based biometrics has emerged as a promising complementary modality for identity verification, particularly in scenarios where the lower facial region is occluded. Early approaches relied on handcrafted features, such as texture descriptors and edge-based representations, to model wrinkle patterns and skin structure. While these methods demonstrated that forehead creases contain identity-discriminative information, they were limited in handling variations in illumination, facial expression, and image resolution.

With the advancement of deep learning, recent works have shifted towards data-driven approaches for modelling forehead characteristics. Methods such as FHCVS~\cite{bharadwaj2022mobile} and FH-SSTNet~\cite{sharma2024fh} introduced specialised architectures designed to capture the spatial arrangement and structural properties of forehead creases. These models explicitly learn representations based on line patterns, curvature, and local texture variations, achieving improved performance over handcrafted pipelines. Similarly, convolutional neural network–based approaches trained directly on forehead regions have demonstrated the ability to extract discriminative features under controlled settings~\cite{bharadwaj2022mobile,sharma2024fh}.
To address the limited availability of annotated datasets, synthetic data generation has also been explored. Synthetic forehead datasets attempt to model variations in crease patterns, shape, and illumination, enabling controlled experimentation and data augmentation. However, models trained on synthetic data often suffer from domain shift when applied to real-world images, resulting in degraded performance.

Despite these advancements, several limitations remain. Most existing approaches are highly task-specific and do not leverage transferable representations from large-scale face recognition models. In addition, evaluations are often conducted on relatively small datasets, limiting insights into generalisation. Furthermore, comparisons across methods are typically restricted, making it difficult to establish a comprehensive understanding of performance across different paradigms. These challenges motivate the need for a more unified and scalable approach that combines domain-specific modelling with transferable feature learning.
\vspace{-0.1in}
\subsection{Face Recognition and Domain Adaptation}
\vspace{-0.1in}
Face recognition has achieved significant success through deep metric learning frameworks trained on large-scale datasets~\cite{schroff2015facenet}. Models such as FaceNet learn embedding spaces where identity-related features are well separated, enabling robust verification across varying conditions. A key strength of these models lies in their ability to learn hierarchical representations that capture both global facial structure and fine-grained texture information.
This property suggests that pretrained face recognition models may be transferable to localised regions such as the forehead. Recent studies have explored domain adaptation strategies, including fine-tuning pretrained networks and employing margin-based loss functions such as AdaFace~\cite{kim2022adaface} and CosFace~\cite{wang2018cosface}. These approaches improve feature discrimination by enforcing intra-class compactness and inter-class separability, particularly under subtle appearance variations.
However, forehead-crease biometrics remain largely underexplored. Existing methods rarely investigate the transferability of pretrained face-recognition representations to this fine-grained and spatially constrained trait, leaving the effectiveness of domain adaptation for forehead-crease verification an open question.

\subsection{Multimodal Large Language Models}

Recent advances in multimodal learning have led to the development of powerful vision-language models that jointly process visual and textual information. These models combine visual encoders with large language models, enabling them to interpret images and generate descriptive or reasoning-based outputs. They have demonstrated strong performance across tasks such as visual question answering, image captioning, and visual reasoning.
A key advantage of these models is their ability to perform zero-shot inference, allowing them to generalise to new tasks without task-specific training. This makes them particularly appealing for biometric applications where labelled data is limited, such as forehead-crease recognition. Moreover, multimodal models offer a unique advantage in terms of explainability. Unlike traditional biometric systems that provide only similarity scores, these models can generate human-interpretable justifications based on visual evidence, enhancing transparency and trust.
Despite these benefits, multimodal models face several limitations in fine-grained biometric tasks. Forehead-crease recognition requires precise analysis of subtle texture patterns and structural variations, which general-purpose models are not explicitly designed to capture. Their performance is sensitive to variations in illumination, resolution, and expression, and they may produce inconsistent predictions or hallucinated reasoning when differences between samples are subtle.
FMs and MLLMs are pretrained on large-scale multimodal corpora and exhibit strong zero-shot and few-shot generalization capabilities. These properties make them attractive for biometric applications, where acquiring large annotated datasets is often difficult. Unlike conventional biometric systems that require task-specific training, MLLMs can perform verification through prompt-driven reasoning and provide human-interpretable explanations. 
However, biometric recognition operates under stringent security constraints, requiring extremely low EER and high TMR at low FMR operating points. Generic foundation models are optimized for broad visual-semantic understanding rather than fine-grained biometric discrimination. Furthermore, large-scale deployment introduces practical challenges related to computational cost, inference latency, and scalability. These limitations motivate a systematic comparison between generic foundation models, domain-adaptive learning, and trait-specific biometric models.

\subsection{Limitations of Existing Work}
Despite recent progress, forehead-crease biometrics remain underexplored, with limited use of pretrained face-recognition models, inconsistent evaluation protocols, and insufficient study of explainability and domain shift.
To address these gaps, we systematically evaluate foundation, domain-adaptive face-recognition, and trait-specific models under a unified protocol, comparing their interpretability, transferability, and verification performance.
\vspace{-0.1in}
\section{Proposed Method}
\label{sec:method}
\vspace{-0.1in}

In this section, we present the proposed unified framework for forehead-crease–based biometric recognition. The goal is to design a reliable verification system that operates solely on the forehead region, particularly in scenarios where full-face information is unavailable due to occlusion or masking. 
Unlike conventional approaches that rely on a single paradigm, our framework integrates three complementary components: 
\textbf{(i)} an explainable \textbf{Zero-Shot} verification module using multimodal reasoning, 
\textbf{(ii)} a \textbf{Domain-adaptive} learning module that transfers knowledge from pretrained face recognition models, and 
\textbf{(iii)} a \textbf{Trait-Specific} model designed to capture fine-grained crease structures. 
The overall pipeline is illustrated in Figure~\ref{main_fig}.

\subsection{Problem Formulation}

Given a pair of forehead images $(\mathbf{x}_i, \mathbf{x}_j)$, the objective is to determine whether they belong to the same identity. This is formulated as a binary verification task:
\begin{equation}
y_{ij} =
\begin{cases}
1, & \text{if } \mathbf{x}_i \text{ and } \mathbf{x}_j \text{ belong to the same subject}, \\
0, & \text{otherwise}.
\end{cases}
\end{equation}

The system produces a similarity score $s_{ij} \in [0,1]$. For explainable models, it additionally outputs a structured textual explanation describing the reasoning behind the decision.
\vspace{-0.1in}
\subsection{Forehead Region Extraction and Preprocessing}
\vspace{-0.1in}
Face images are first processed to extract the forehead region using a fixed cropping strategy, ensuring spatial consistency across samples. As forehead creases are predominantly structural and textural cues, the cropped regions are converted to grayscale to reduce appearance variability. The resulting images are replicated across three channels and resized to $224 \times 224$ for compatibility with standard CNN architectures. During training, horizontal flipping, small rotations, and brightness perturbations are applied to improve generalization while preserving discriminative crease patterns.

\begin{figure*}[!ht]
\centering
\includegraphics[width=\textwidth]{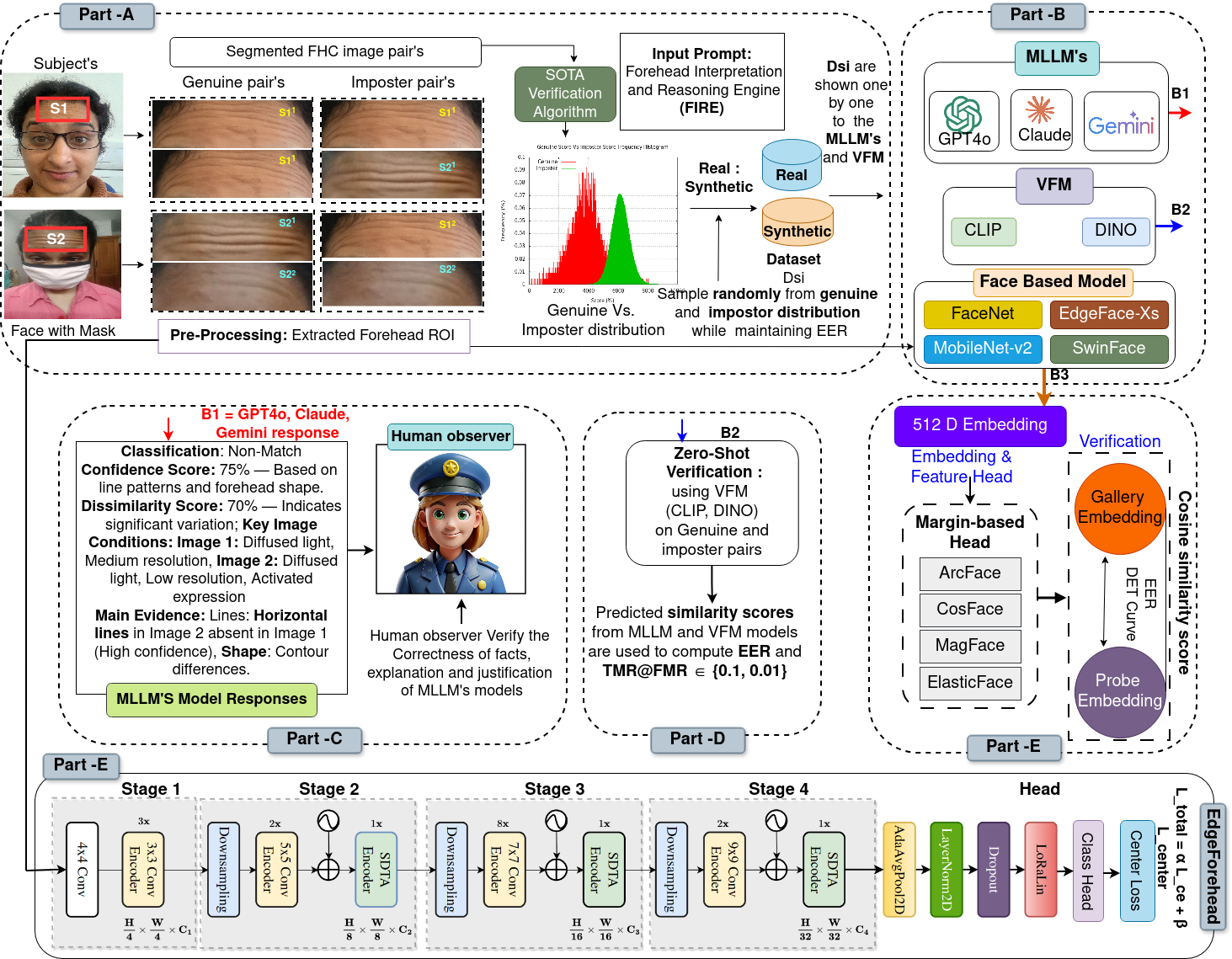}
\vspace{-2pt}
\caption{Forehead-crease verification pipeline using real and synthetic image pairs evaluated with MLLMs, FMs and trait-specific models.}
\label{main_fig}
\end{figure*}

\vspace{-0.1in}
\subsection{Explainable Zero-Shot Verification via FIRE}
\vspace{-0.1in}
To enable interpretable biometric verification, we introduce the \textbf{Forehead Interpretation and Reasoning Engine (FIRE}(detailed prompt in supplementary material), a structured prompting framework that guides multimodal large language models (MLLMs) such as Claude~\cite{claude2025}, GPT-4o~\cite{gpt2024}, and Gemini~\cite{team2023gemini}.
Unlike supervised models, these systems are not explicitly trained for biometric verification. Therefore, we formulate the task as a structured reasoning problem using Chain-of-Thought (CoT) prompting.
Each input consists of a pair of forehead images along with a carefully designed prompt that encourages step-by-step comparison.
The \textbf{FIRE} framework decomposes the verification process into multiple stages, including: \textbf{(i)} assessment of image conditions (lighting, occlusion, resolution),\textbf{ (ii)} extraction of structural features (crease patterns, spacing, orientation), \textbf{(iii)} comparison of primary and secondary features, \textbf{(iv)} decision-making with confidence and dissimilarity estimation.
This structured reasoning reduces hallucinations and improves consistency compared to generic prompts. The model outputs both a classification (match/non-match) and a human-readable explanation, enabling transparent and interpretable verification. The complete FIRE (detailed prompt provided in the supplementary).

\vspace{-0.1in}
\subsection{Zero-Shot Verification using Vision Foundation Models}
\vspace{-0.1in}
To further analyse zero-shot generalisation, we evaluate vision foundation models (VFMs), including DINOv2~\cite{oquab2023dinov2} and CLIP~\cite{radford2021learning}, without task-specific fine-tuning.
Given an input forehead image $\mathbf{x}_i$, an embedding $\mathbf{z}_i$ is extracted using a pretrained model. Verification is performed using cosine similarity:
\begin{equation}
s_{ij} = \frac{\mathbf{z}_i \cdot \mathbf{z}_j}{\|\mathbf{z}_i\| \|\mathbf{z}_j\|}
\end{equation}
DINOv2 captures fine-grained structural features due to self-supervised training, while CLIP benefits from semantic alignment learned from image-text pairs. These models serve as strong zero-shot baselines for evaluating generalisation beyond traditional face recognition cues.
However, since these models are not trained specifically for forehead biometrics, their performance is sensitive to variations in illumination, expression, and domain shift, motivating the need for domain adaptation.
\subsection{Domain-Adaptive Learning using Pretrained Models}
To address the limitations of zero-shot models, we adopt a domain-adaptive learning strategy that transfers knowledge from pretrained face recognition models to the forehead domain.
\subsubsection{Feature Representation}
Let $f(\cdot)$ denote a pretrained face recognition network. For an input image $\mathbf{x}_i$, the model produces an embedding:
$\mathbf{z}_i = f(\mathbf{x}_i), \quad \mathbf{z}_i \in \mathbb{R}^{512}$
These embeddings capture identity-related information, which is further refined for forehead-specific features through fine-tuning. We employ multiple backbone architectures, including  MobileFaceNet-V2~\cite{chen2018mobilefacenets}, EdgeFace-Xs~\cite{10388036}, SwinFace~\cite{qin2023swinface}, and FaceNet~\cite{schroff2015facenet}.
\subsubsection{Margin-Based Metric Learning}

To enhance discriminability, we employ margin-based loss functions that enforce compact intra-class distributions and well-separated inter-class boundaries.

We further evaluate several margin-based objectives, including \textbf{ElasticFace} (stochastic margin regularization)  ~\cite{boutros2022elasticface}, \textbf{AdaFace}~\cite{kim2022adaface} (quality-adaptive margin), \textbf{MagFace}~\cite{meng2021magface} (magnitude-aware representation learning) , and  \textbf{CosFace}~\cite{wang2018cosface} (additive cosine margin) . These objectives enhance intra-class compactness and inter-class separability, enabling more discriminative modeling of fine-grained forehead-crease patterns for robust verification.
\begin{tcolorbox}[
    colback=gray!4,
    colframe=blue!30!white,
    title=\textbf{Forehead Interpretation and Reasoning Engine (FIRE)},
    coltitle=black,
    fonttitle=\bfseries,
    boxrule=0.5pt,
    arc=3mm,
    width=\textwidth,
    before skip=2pt,
    after skip=2pt,
    boxsep=1pt,
    left=2pt,
    right=2pt,
    top=1pt,
    bottom=1pt
]
\small
\textbf{Input:} Two forehead images.
\textbf{Output:} Match/Non-Match, Confidence Score, Dissimilarity Score, and Explanation.
\textbf{Step 1:} Assess image quality (lighting, occlusion, resolution, expression).
\textbf{Step 2:} Extract and compare forehead-crease characteristics, including topology, spacing, orientation, continuity, forehead shape, and permanent marks.
\textbf{Step 3:} Normalize illumination, scale, occlusion, and expression variations.
\textbf{Step 4:} Evaluate feature consistency, assign confidence weights, and estimate similarity.
\textbf{Step 5:} Generate the final decision with confidence, dissimilarity score, and reasoning.
\textbf{Constraint:} Use only observable forehead features and explicitly report uncertainty when evidence is insufficient.
\end{tcolorbox}
\subsection{Trait-Specific Model: Edge-Forehead}

While domain-adaptive models improve performance, they are still derived from general face representations. To further enhance fine-grained discrimination, we introduce a trait-specific model, \textbf{Edge-Forehead}, tailored for forehead-crease recognition.
We fine-tune the EdgeFace model~\cite{10388036} on forehead images using a gradual unfreezing strategy, allowing the network to retain general facial features while progressively specialising in crease-specific patterns.
To improve embedding quality, we employ a dual-loss objective:
\begin{equation}
\mathcal{L}_{\text{total}} = \alpha \mathcal{L}_{\text{ce}} + \beta \mathcal{L}_{\text{center}}
\end{equation}

where $\mathcal{L}_{\text{ce}}$ enforces inter-class separability and $\mathcal{L}_{\text{center}}$ reduces intra-class variance. The weighting parameters $\alpha$ and $\beta$ balance classification accuracy and feature compactness.
By focusing exclusively on forehead regions, the model(Figure\ref{main_fig}, Part E) learns fine-grained structural characteristics such as crease continuity, spacing, curvature, and texture patterns. This leads to improved robustness under variations in illumination, pose, and expression.
Finally, all models are evaluated under a unified verification protocol, enabling a fair comparison between zero-shot, domain-adaptive, and trait-specific approaches.

\vspace{-0.1in}
\section{Dataset \& Evaluation Protocol}
\vspace{-0.1in}
\label{sec:dataset_and_eval_protocol}

In this section, we describe the datasets and evaluation protocol used to assess the proposed framework across zero-shot, domain-adaptive, and trait-specific models. Forehead crease patterns are considered as the primary biometric trait. Experiments are conducted on both real and synthetic datasets to ensure comprehensive evaluation under diverse conditions. We design a unified protocol to enable fair comparison across MLLMs, VFMs, and supervised forehead-crease recognition systems.

\subsection{Datasets for Zero-Shot Evaluation (LLMs \& VFMs)}
\label{sec:dataset_llm}
We evaluate zero-shot verification on two benchmark datasets subsets: a real dataset (FH-V2) and a synthetic dataset~\cite{tandon2024synthetic}. The synthetic dataset was obtained from~\cite{tandon2024synthetic} and used to evaluate forehead verification under synthetic image conditions. Masked face images are first processed to extract the forehead region, as shown in Figure ~\ref{main_fig}. To construct a reliable benchmark, a pretrained SOTA verification model~\cite{tandon2024synthetic} is used to compute similarity scores and select representative genuine and impostor pairs. Specifically, 50 genuine pairs from 50 unique subjects and 450 impostor pairs from 250 distinct subjects are sampled from both datasets. This high negative-to-positive ratio follows standard biometric evaluation protocols, ensuring robust score separation and stable performance estimation. The selected pairs capture variations in illumination, expression, and image quality, enabling a realistic assessment of MLLMs and VFMs for forehead-crease-based verification.



\subsection{Datasets for Domain-Adaptive \& Trait-Specific Models}
\label{sec:dataset_supervised}
For supervised training, we use the FH-V1 dataset~\cite{bharadwaj2022mobile}, obtained by extracting forehead regions from face images using a fixed YOLOv6\cite{li2022yolov6} based cropping strategy. The dataset comprises 247 subjects with  2,462 gallery images and 2,502 probe images, totaling 4,964 images and variations in illumination, pose, expression, and image quality. Subjects are treated as unique identity classes, and a subject-disjoint train-validation split is employed to prevent identity leakage and support model selection. For open-set evaluation, all models are tested on FH-V2~\cite{tandon2024synthetic}, comprising 295 unseen subjects and 29,500 genuine and approximately 8.67 million impostor verification pairs.
\vspace{-0.15in}
\subsection{Training Protocols}
Domain-adaptive and trait-specific models are trained on FH-V1 using a stratified 90\%/10\% train-validation split. Pretrained face-recognition backbones are fine-tuned using margin-based objectives (CosFace, AdaFace, MagFace, and ElasticFace). Edge-Forehead is initialized with EdgeFace-Base weights and trained in three stages: (i) cross-entropy with label smoothing ($\epsilon=0.1$) and progressive unfreezing (70\%,$\rightarrow$,0\%), (ii) joint cross-entropy and center loss ($\lambda_{center}=0.3$, $\alpha=0.5$) with augmentation (flip, $\pm9^\circ$ rotation, 0.8--1.2$\times$ scaling, grayscale), and (iii) semi-hard triplet mining with a dynamic margin (0.4--1.5). Stages 1--2 use SGD (lr=$10^{-3}$, momentum=0.9, weight decay=$5\times10^{-4}$, batch size=32), while Stage 3 uses Adam (lr=$10^{-5}$, batch size=2). Early stopping (patience=3) is applied throughout training. 
Images are resized to $112\times112$. Verification uses Euclidean distance between 512-D embeddings, with thresholds set at EER. We use EdgeFace-Base (18.23M parameters) with dropout ($p=0.5$) and a 247-class FC head. Training uses three stages: progressive unfreezing (70\%, 50\%, 30\%, 10\%, and 0\% frozen; up to 15 epochs per setting), Stage 2 (up to 30 epochs), and Stage 3 (until convergence). The center-update rate and center-loss weight are $\alpha=0.5$ and $\beta=0.3$, respectively.


\subsection{Evaluation Protocol}
To comprehensively evaluate the proposed framework, we consider three categories of models: (i) Multimodal LLMs: GPT-4o, Claude Sonnet 4, and Gemini 2.5 Pro. (ii) Vision foundation models: DINOv2 and CLIP (iii) Supervised models: Edge-Forehead (proposed), and FHCVS.
All models are evaluated under a unified open-set verification protocol on both real (FH-V2) and synthetic datasets.
For \textbf{zero-shot models} (MLLMs and VFMs), each image pair undergoes three independent stateless inference runs to reduce output stochasticity. Final decisions are obtained via majority voting, with similarity/confidence scores averaged across runs. When refusals or conflicting predictions occur, the highest-confidence prediction is retained.
For \textbf{domain-adaptive} and \textbf{trait-specific} models, verification is performed using cosine similarity in the learned embedding space. Scores are computed and aggregated uniformly across models to ensure fair comparison. Performance is evaluated using standard biometric metrics, including Equal Error Rate (EER) and True Match Rate (TMR) at fixed False Match Rate (FMR) operating points of 0.1 and 0.01. These metrics provide a consistent framework for comparing zero-shot, domain-adaptive, and trait-specific approaches under varying operating conditions.

\textbf{Reproducibility Details:} We use a subject-disjoint 90\%/10\% train validation split(seed=42). Verification pairs follow Section~4.1. Images are resized to $112\times112$, normalized, and augmented. Edge-Forehead is initialized from pretrained EdgeFace-Base and trained using the three-stage schedule in Section~4.3. FIRE prompts, model versions, and three-run stateless inference settings are fixed. Verification thresholds are selected at EER using Euclidean distances between 512-D embeddings. It has approximately 18.23M parameters and requires 1,398.83 MFLOPs per forward pass.

\vspace{-0.1in}
\section{Results \& Discussion}
\vspace{-0.1in}
\label{result_discussion}
This section evaluates whether generic foundation models (GPT-4o, Gemini 2.5 Pro , Claude Sonnet 4) can provide competitive verification performance to domain-adaptive and trait-specific biometric systems. While foundation models offer zero-shot generalization and explainability, biometric verification demands highly discriminative representations and strict operating characteristics. The experiments analyse this trade-off across all three paradigms under a unified protocol.
\subsection{FIRE-based LLM and VFM Results}
We first evaluate the performance of FIRE-based multimodal LLMs and vision FMs on both real and synthetic datasets for zero-shot verification. The quantitative results are summarised in Table~\ref{table:fire_results} and explainability examples in supplementary material Figure 3.
\subsubsection{Quantitative Analysis}
\begin{table}[!t]
\centering
\scriptsize
\renewcommand{\arraystretch}{1.1}
\setlength{\tabcolsep}{4pt}
\caption{Identity verification results on the sampled FH-V2 verification subset (50 genuine and 450 impostor pairs) and synthetic subset.
\cite{tandon2024synthetic}.}
\begin{tabular}{lccc}
\hline
\textbf{Model} & \textbf{EER (\%) $\downarrow$} & \textbf{TMR@FMR} & \textbf{TMR@FMR} \\
\textbf{} & \textbf{  } & \textbf{=0.1 (\%) $\uparrow$} & \textbf{=0.01 (\%) $\uparrow$} \\
\hline
\multicolumn{4}{c}{\textbf{FH-V2(Real}} \\ \hline
\textbf{Edge-Forehead(Our)} & \textbf{8.63}  & \textbf{70.00} & \textbf{70.00} \\ 
FH-SSTNet\cite{sharma2024fh} & 14.22 & 38.00 & 38.00 \\
FHCVS \cite{tandon2024synthetic} & 21.56 & 32.00 & 32.00 \\
CLIP \cite{radford2021learning} & 24.11 & 14.00 & 6.00 \\
DINO-V2 \cite{oquab2023dinov2} & 24.00 & 40.00 & 38.00 \\
Claude Sonnet 4\cite{claude2025} & 27.11 & 24.00 & 24.00 \\
Gemini 2.5 Pro \cite{team2023gemini} & 24.89 & 26.00 & 26.00 \\
GPT-4o \cite{gpt2024} & 18.22 & 52.00 & 52.00 \\
\hline 
\multicolumn{4}{c}{\textbf{Synthetic}} \\ \hline
\textbf{Edge-Forehead(Our)} & \textbf{0.25} & \textbf{94.00} & \textbf{94.00} \\
FH-SSTNet\cite{sharma2024fh} & 29.74 & 5.88 & 3.92 \\
FHCVS \cite{tandon2024synthetic} & 24.22 & 22.00 & 16.00 \\
CLIP \cite{radford2021learning} & 31.56 & 10.00 & 4.00 \\
DINO-V2 \cite{oquab2023dinov2} & 28.00 & 12.00 & 12.00 \\
Claude Sonnet 4\cite{claude2025} & 31.89 & 18.00 & 18.00 \\
Gemini 2.5 Pro\cite{team2023gemini} & 30.89 & 22.00 & 10.00 \\
GPT-4o \cite{gpt2024} & 25.44 & 20.00 & 12.00 \\
\hline
\end{tabular}
\label{table:fire_results}
\end{table}
For reference, the trait-specific Edge-Forehead model achieves the best performance on the sampled FH-V2 evaluation subset (EER = 8.63\%), while its full open-set evaluation on FH-V2 is reported separately in Section~5.2.
Results on the complete FH-V2 open-set evaluation are reported separately in Table \ref{tab:verification}, where Edge-Forehead attains an EER of 8.30\% with a True Match Rate (TMR) of 70.00\% at both operating points. This represents a significant improvement over existing supervised approaches such as FHCVS (EER: 21.56\%) and FH-SSTNet (EER: 14.22\%).
Among zero-shot methods, GPT-4o achieves the strongest performance (EER: 18.22\%, TMR: 52.00\%), indicating that structured reasoning enables effective capture of identity-related cues even without task-specific training. DINOv2 also demonstrates competitive performance (TMR: 40.00\%, 38.00\%), suggesting that self-supervised representations retain sensitivity to fine-grained structural patterns. In contrast, CLIP exhibits relatively weaker performance, likely due to its emphasis on semantic alignment rather than fine texture discrimination.
On synthetic data, most zero-shot models degrade under domain shift, whereas Edge-Forehead maintains strong performance (EER: 0.25\%; TMR: 94.00\%). GPT-4o performs best among zero-shot models (EER: 25.44\%) but struggles with synthetic variations. Overall, trait-specific learning improves verification accuracy and robustness, while zero-shot models offer interpretability.
\vspace{-0.12in}
\subsubsection{Qualitative Analysis}
Table~\ref{table:qualitative_results} presents a three-round response analysis to evaluate the consistency and reliability of FIRE-based verification. Several key behavioural patterns can be observed:
(i) \textbf{Progressive Refinement}, where predictions often improve across rounds; (ii) \textbf{High-Confidence Stability}, where clear samples yield consistent decisions; (iii) \textbf{Recovery from Refusal}, where initial refusals may be resolved through subsequent prompting; and (iv) \textbf{Response Variability}, where challenging cases can produce conflicting predictions due to model sensitivity and stochasticity. Additional prompt-ablation and stability analyses are provided in the supplementary material.
\vspace{-0.2in}
\begin{table*}[!ht]
\centering
\scriptsize
\renewcommand{\arraystretch}{0.9}
\setlength{\tabcolsep}{2pt}
\resizebox{\textwidth}{!}{%
\begin{tabular}{|p{2.8cm}|p{2.8cm}|p{2.8cm}|p{2.8cm}|}
\hline
\textbf{Scenario} & \textbf{Round 1} & \textbf{Round 2} & \textbf{Round 3} \\
\hline

\textbf{Improved classification results and confidence}
& CLS: \textcolor{red}{Non-Match} \newline Conf: 55\% \newline Dis Score: 65\%
& CLS: \textcolor{green}{Match} \newline Conf: \textbf{70\%} \newline Dis Score: 15\%
& CLS: \textcolor{green}{Match} \newline Conf: \textbf{85\%} \newline Dis Score: 20\% \\
\hline

\textbf{Improved match consistency with lower dissimilarity scores}
& CLS: Match \newline Conf: 85\% \newline Dis Score: \textbf{15\%}
& CLS: Match \newline Conf: 85\% \newline Dis Score: \textbf{15\%}
& CLS: Match \newline Conf: 85\% \newline Dis Score: \textbf{10\%} \\
\hline

\textbf{Improved forehead verification after iterative prompting}
& \textcolor{red}{Unable to perform forensic analysis}
& CLS: NM \newline Conf: 60\% \newline Dis Score: 75\%
& CLS: NM \newline Conf: 75\% \newline Dis Score: 80\% \\
\hline

\textbf{Ambiguity due to conflicting or missing classifications}
& CLS: \textcolor{green}{Match} \newline Conf: 75\% \newline Dis Score: 25\%
& \textcolor{red}{Unable to perform forensic analysis or classify}
& CLS: \textcolor{red}{NM} \newline Conf: 70\% \newline Dis Score: 60\% \\
\hline
\end{tabular}%
}
\caption{Round-wise response comparison across three inference passes. CLS: classification; Conf: confidence; Dis: dissimilarity; M: match; NM: non-match.}
\label{table:qualitative_results}
\end{table*}

\vspace{-0.6in}
\subsubsection{Explainability Analysis}
make it crisp  but keep important information 
A key advantage of FIRE framework is its ability to provide human-interpretable explanations alongside verification decisions. Representative examples of genuine and impostor pairs are presented in the supplementary material with model-generated reasoning. For impostor pairs, FIRE correctly identifies differences in crease alignment and spacing despite illumination and resolution variations, achieving up to 85\% confidence. For genuine pairs, it recognizes consistent crease patterns, producing high-confidence matches (95\%) with low dissimilarity. These results demonstrate that FIRE enhances interpretability while generally aligning with human reasoning, although occasional errors remain under challenging visual conditions.
\vspace{-0.1in}
\subsection{Domain-Adaptive \& Trait specific Forehead-Crease Verification}

\subsubsection{Quantitative Results}
Table~\ref{tab:verification} reports open-set verification performance for domain-adaptive models trained on FH-V1 and evaluated on FH-V2. The results reveal consistent trends across backbone architectures and margin-based loss functions.
\begin{table}[!ht]
    \centering
    \scriptsize
    \renewcommand{\arraystretch}{1.1}
    \setlength{\tabcolsep}{4pt}
    \caption{Open-set verification performance of foundation, domain-adaptive, face-recognition, and trait-specific models on FH-V2 after fine-tuning on FH-V1.}
    \begin{tabular}{lcccc}
    \hline
    \multirow{2}{*}{\textbf{Model}} & \multirow{2}{*}{\textbf{EER (\%)}} & \multicolumn{2}{c}{\textbf{TMR (\%) @ FMR (\%) =}} \\ \cline{3-4}
     &  & \textbf{0.1 (\%)} & \textbf{0.01 (\%)} \\ \hline

FaceNet\cite{schroff2015facenet} + Arcface  & 20.65 & 11.72 & 2.49 \\
FaceNet\cite{schroff2015facenet} + AdaFace  & 15.38  & 26.13 & 8.07 \\
FaceNet\cite{schroff2015facenet} + CosFace  &  17.35 & 17.51 & 4.14 \\
FaceNet\cite{schroff2015facenet} + MagFace  & 26.73 & 7.33 &  1.56\\
FaceNet\cite{schroff2015facenet} + ElasticFace & \textbf{11.17} & \textbf{45.48} & \textbf{27.87} \\
\hline
MFN-V2\cite{chen2018mobilefacenets} + Arcface & 10.73 & 55.76 &  41.04\\
MFN-V2\cite{chen2018mobilefacenets} + AdaFace & 13.68 & 46.35 & 32.32 \\
MFN-V2\cite{chen2018mobilefacenets} + CosFace & \textbf{9.31} & \textbf{60.26} &  \textbf{44.59}\\
MFN-V2\cite{chen2018mobilefacenets} + MagFace & 11.51 & 51.00 & 32.22 \\
MFN-V2\cite{chen2018mobilefacenets} + ElasticFace & 15.37 & 41.91 & 26.33 \\
\hline
SwinFace\cite{qin2023swinface} + Arcface  & 17.72 & 28.61 &  \textbf{10.12} \\
SwinFace\cite{qin2023swinface} + AdaFace  & \textbf{13.44} & \textbf{31.64} & 9.43 \\
SwinFace\cite{qin2023swinface} + CosFace  &  21.33 & 21.50 & 7.55 \\
SwinFace\cite{qin2023swinface} + MagFace  & 22.13 & 6.80 & 1.93 \\
SwinFace\cite{qin2023swinface} + ElasticFace & 22.13 & 14.80 & 4.82 \\
\hline
EF-Xs\cite{10388036} + Arcface  & 21.54 & 24.14 &  7.73 \\  
EF-Xs\cite{10388036} + AdaFace  & 12.72 & 49.01 &  33.83\\
EF-Xs\cite{10388036} + CosFace  & 19.46 & 27.47 & 11.52 \\
EF-Xs\cite{10388036} + MagFace  & \textbf{11.01} & \textbf{52.14} & \textbf{37.35} \\
EF-Xs\cite{10388036} + ElasticFace & 22.81 & 22.12 & 6.43 \\
\hline
DINO-V2 (ViT-b)\cite{oquab2023dinov2} & 8.83  & \textbf{57.16}  & \textbf{40.39} \\  
CLIP(ViT-b)\cite{radford2021learning} & 10.72 & 44.25 & 25.29  \\
FHCVS\cite{tandon2025generating}   & 12.37  & 47.61 & 35.93   \\
ResNet 18 + CAM\cite{bharadwaj2022mobile}  & 14.61  & 47.61  & 35.93   \\
EdgeFace-Base\cite{10388036}  & 9.84 & 51.95 & 35.12  \\ 
Edgeface-Base\cite{10388036}(inference) & 26.89 & 8.50 & 3.00 &  \\ 
Edgeface-Base\cite{10388036} + CosFace & 18.15 & 38.97 & 25.19 &  \\ 
Edgeface-Base\cite{10388036} + Adaface & 20.95 & 36.75 & 22.80 &  \\
\textbf{EdgeForehead(Our)}  & \textbf{8.30} & 54.01 & 37.08 &   \\ 

\hline

    \end{tabular}    
    \label{tab:verification}
\end{table}
For the FaceNet backbone, margin-based supervision significantly improves performance compared to standard formulations. ElasticFace achieves the best results (EER: 11.17\%), benefiting from stochastic margin sampling that improves generalisation under limited identity evidence.
MobileFaceNet-V2 demonstrates the strongest overall performance across all architectures. In particular, MobileFaceNet-V2 combined with CosFace achieves the lowest EER of 9.31\% and the highest TMR values. This can be attributed to its lightweight convolutional design, which effectively captures local texture patterns critical for forehead-based recognition.
SwinFace exhibits comparatively lower performance, with AdaFace achieving the best results within this group (EER: 13.44\%). The reduced performance suggests that global attention mechanisms may be less effective for modelling highly localised crease structures.
For EdgeFace-Xs, MagFace achieves the best performance (EER: 11.01\%), highlighting the importance of quality-aware feature learning in handling variations in image quality and crease visibility.
These findings indicate that both backbone architecture and margin-based supervision play a crucial role in capturing fine-grained biometric features.

\begin{figure}[!t]
\centering
\includegraphics[width=0.7\textwidth]{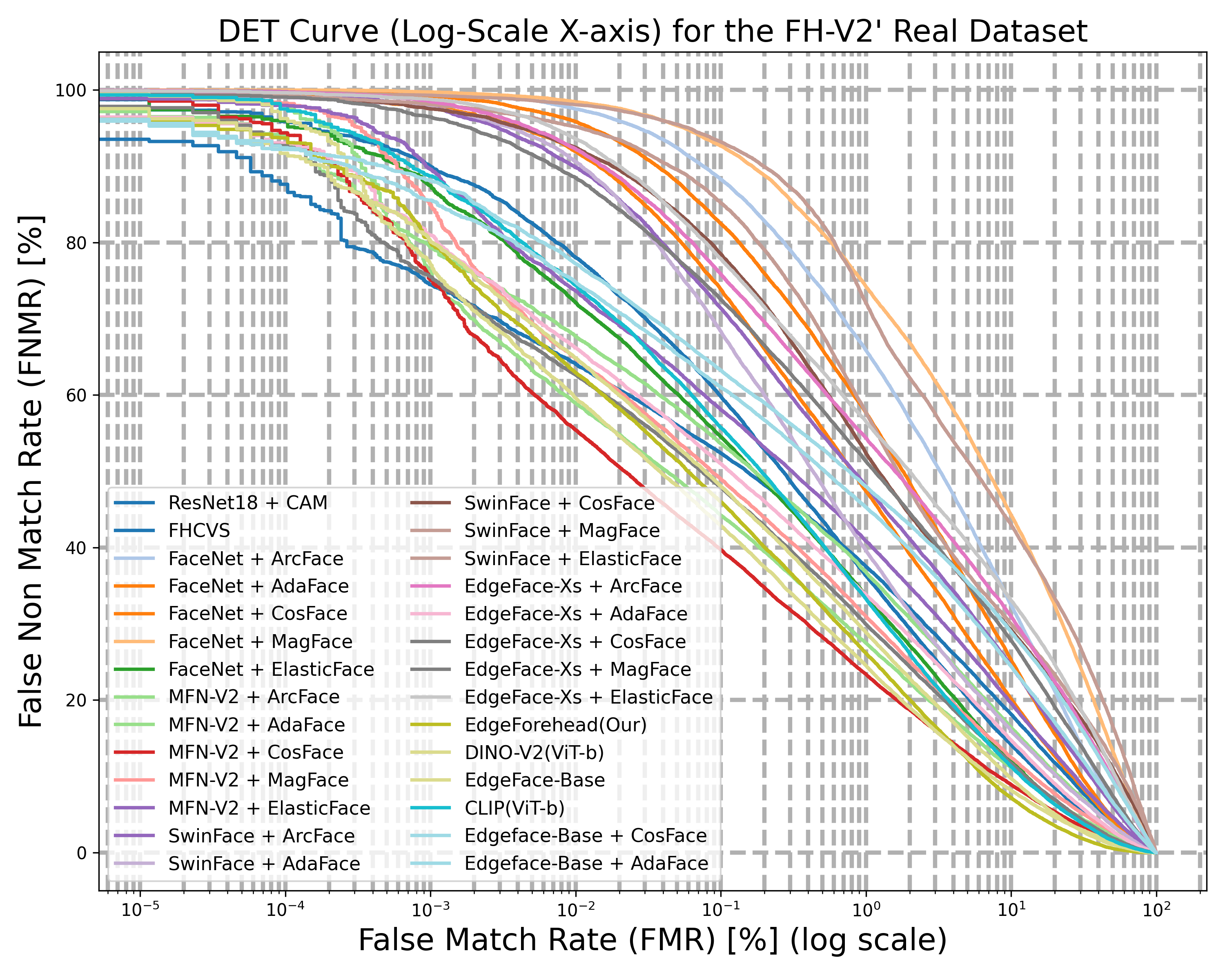}
\caption{DET curves for forehead-crease verification on the open-set FH-V2 dataset.}
\label{fig:det_curve}
\end{figure}
\textbf{Trait-Specific Forehead-Crease Verification}: Table~\ref{tab:verification} compares Edge-Forehead with existing forehead-recognition approaches. Edge-Forehead achieves the lowest EER (8.30\%) and competitive TMRs, surpassing Vision Foundation Models (DINOv2, CLIP). The results demonstrate that explicit modeling of fine-grained crease structures yields superior discriminability and robustness, underscoring the effectiveness of trait-specific learning for forehead-crease verification. EdgeFace-Base is fine-tuned on FH-V1, while EdgeFace-Base (inference) uses the original pretrained model without fine-tuning.
Figure~\ref{fig:det_curve} shows the \textbf{DET curves} for all models. EdgeForehead achieves the best overall trade-off, attaining the lowest EER (8.30\%), while MFN-V2+CosFace delivers the strongest low-FMR performance. SwinFace variants exhibit higher FNMRs, suggesting weaker modeling of fine-grained crease patterns. These results highlight the effectiveness of locality-preserving architectures and trait-specific learning for robust forehead-crease verification.
\vspace{-0.1in}
\section{Conclusion}
\vspace{-0.1in}
\label{conclusion}
This work presents a systematic comparison of MLLMs (GPT-4o, Claude, Gemini), domain-adaptive learning, and trait-specific models for forehead-crease biometric verification. While MLLMs enable interpretable zero-shot verification, their performance remains below specialized biometric systems. Domain-adaptive and trait-specific models consistently achieve lower EER and higher TMR, highlighting the importance of discriminative biometric learning. These findings indicate that FMs offer interpretable verification and valuable zero-shot capabilities, but security-critical biometric applications still require trait-specific learning for reliable deployment.

\bibliographystyle{splncs04}
\bibliography{ref}

\end{document}


\title{Supplementary Material: Balancing Foundation Models and Trait-Specific Learning for Forehead-Crease Recognition}
\author{Geetanjali Sharma\inst{1} \and
Aryan Jain\inst {1} \and
Dev Nagaich\inst{1} \and
Aditya Nigam\inst{1} \and
Raghavendra Ramachandra \inst{2}}
\authorrunning{Geetanjali Sharma et al.}
%
\institute{Indian Institute of Technology Mandi, India \and
SAFE Center, Norwegian University of Science and Technology (NTNU), Gjøvik, Norway
\\}

%
%
%
%
\maketitle              
\section{Detailed FIRE Prompt and Reasoning Protocol)}

The Forehead Interpretation and Reasoning Engine (FIRE) is a structured prompting framework designed to guide multimodal foundation models towards consistent and explainable forehead-crease verification. Given a pair of forehead images, FIRE produces a Match/Non-Match decision, confidence score, dissimilarity score, and a structured justification.

FIRE consists of five stages: (i) image quality assessment, including illumination, occlusion, resolution, and expression analysis; (ii) extraction of discriminative forehead characteristics such as crease topology, spacing, continuity, orientation, forehead shape, and visible permanent marks; (iii) normalization for illumination, scale, occlusion, and expression variations; (iv) evidence aggregation through confidence-weighted feature comparison; and (v) decision generation with a confidence score, dissimilarity score, and interpretable explanation.

The framework explicitly restricts reasoning to observable visual evidence and requires uncertainty reporting when image quality is insufficient. By enforcing a structured feature-level analysis, FIRE reduces hallucinations and improves the reliability of zero-shot biometric verification.

\definecolor{promptbg}{RGB}{240,248,255}  
\definecolor{promptborder}{RGB}{0,102,204} 
\begin{tcolorbox}[
    colback=gray!4,
    colframe=blue!30!white,
    title=\textbf{Forehead Interpretation and Reasoning Engine (FIRE)},
    coltitle=black,
    fonttitle=\bfseries,
    boxrule=0.5pt,
    arc=3mm,
    width=\textwidth,
    breakable
]
\textbf{Input}- Two forehead creases images (Image 1, Image 2).  \textbf{Output - } Match / Non-Match, Confidence Score (\%), Dissimilarity Score (\%), Structured forensic justification \\
\textbf{Step 0:} Assumptions \& Bias Control, \textbf{Initialize Null Hypothesis:} identity relationship = unknown, Decisions based only on observable features, Normalize mentally for lighting, scale, pose, expression. 
\textbf{Step 1:} Image Quality \& Condition Assessment, For each image, record: 
\textbf{Lighting (L):} uniform / uneven / shadowed, \textbf{Occlusion (O):} hair / partial coverage, \textit{Resolution} \& Scale (R/S): relative proportions only \textit{Expression (E):} Activated / Relaxed / Intermediate / Asymmetric, Identify common visible forehead region \& landmarks.
\textbf{Step 2:} Case Assignment (Expression State), \textbf{Case A:} Both Activated, \textbf{Case B:} Mixed (Activated + Relaxed), \textbf{Case C:} Both Relaxed, \textbf{Step 3:} Feature Extraction \& Comparison, \textbf{Primary Features} (High Weight), \textit{Major Wrinkle Line Patterns}, Count, spacing, relative position, Compare using ratios, not absolute distances, Case-dependent adaptation (A/B/C), \textit{Forehead Shape} \& Contour, Height–width ratio,\textit{ curvature}, structural layout, Focus on central shared region only, \textit{Permanent Marks, Scars, moles, stable marks }(if visible), Assign Confidence: H / M / L per feature,\textbf{ Secondary Features} (Supportive Evidence), \textit{Line depth} \& prominence, Bilateral symmetry patterns, Skin texture (resolution-aware), \textit{Line orientation} \& flow direction, Anatomical proportions (brow–hairline ratios), \textbf{Step 4:} Normalization \& Robustness Handling, Lighting normalization, Occlusion-aware comparison, Scale compensation via \textit{wrinkle spacing}, Expression compensation using case logic.
\textbf{Step 5:} Decision Logic, Evaluate primary feature consistency, Validate with secondary feature agreement, Weight features by confidence level, Check inter-feature coherence.
\textbf{Step 6:} Scoring Classification: Match / Non-Match, Confidence Score (\%): reliability of evidence, Dissimilarity Score (\%): degree of feature mismatch, 0\% = identical, 100\% = entirely different,.
\textbf{Step 7:} Reporting (Structured) Image conditions: [L/O/R/E], Key challenges, 
Primary \& secondary findings with confidence, Conflicting evidence (if any), 
Final reasoning + uncertainty acknowledgment.
\textbf{Quality Constraints}: No assumptions beyond visible data, Report uncertainty when evidence is limited, Prioritize structural features over surface noise, Always produce a decision with justification.
\end{tcolorbox}

\section{Ablation study}
\label{ablation_study}


\subsection{Prompt Ablation Study}
\label{ablation1}
To evaluate the impact of prompt design, we compare a Forehead-Aware Short Prompt (SP1) with the proposed FIRE prompt using a subset of FH-V2 consisting of 10 genuine and 90 impostor pairs. Both prompts are evaluated under the same zero-shot protocol using GPT-4o.

Table 1 shows that FIRE consistently outperforms SP1 across both real and synthetic datasets. On FH-V2, FIRE reduces the EER from 25.56\% to 18.33\%, while on the synthetic dataset the EER decreases from 28.33\% to 18.85\%. Similar improvements are observed for TMR at fixed FMR operating points.

These results indicate that structured reasoning plays a critical role in forehead-crease verification. By explicitly guiding the model towards crease topology, spacing, continuity, and texture analysis, FIRE improves both discriminative performance and interpretability.

\begin{tcolorbox}[
    colback=gray!4,
    colframe=blue!30!white,
    title=\textbf{Forehead Aware specific Short Prompt(SP1)},
    coltitle=black,
    fonttitle=\bfseries,
    boxrule=0.5pt,
    arc=2mm,
    width=\textwidth,
    breakable
]
\textbf{SP1:} These are the images of the forehead region. Understand the feature(e.g., creases, lines, skin texture, skin color \& eyebrow status) visible in this image. Please provide a classification (Match or Non-Match) along with dissimilarity scores (0-100\%) and explain your description of the feature based purely on the visual characteristics (such as number, clarity, pattern, density, or contrast), without making assumptions beyond what is shown.
\end{tcolorbox}

Table~\ref{table:abl_results} presents the quantitative comparison. The SP1 prompt results in higher error rates and lower verification performance, particularly on the synthetic dataset. It also exhibits inconsistent responses and increased hallucinations due to the lack of structured reasoning guidance.
\begin{table}[htbp]
\centering
\scriptsize
\renewcommand{\arraystretch}{1.1}
\setlength{\tabcolsep}{4pt}
\begin{tabular}{lccc}
\hline
\textbf{Model} & \textbf{EER (\%) $\downarrow$} & \textbf{TMR@FMR} & \textbf{TMR@FMR} \\
\textbf{} & \textbf{  } & \textbf{=0.1 (\%) $\uparrow$} & \textbf{=0.01 (\%) $\uparrow$} \\
\hline
\multicolumn{4}{c}{\textbf{FH-V2}} \\
GPT-4o (SP1) & 25.56 & 50.00 & 50.00 \\
GPT-4o (FIRE Prompt)  & \textbf{18.33} & \textbf{70.00} & \textbf{70.00} \\
\hline
\multicolumn{4}{c}{\textbf{Synthetic}} \\
GPT-4o (SP1) & 28.33 & 10.00 & 10.00 \\
GPT-4o (FIRE Prompt) & \textbf{18.85} & \textbf{60.00} & \textbf{60.00} \\

\hline
\end{tabular}
\caption{Forehead-crease verification results using SP1 and FIRE prompts with GPT-4o on FH-V2 and synthetic datasets.}
\label{table:abl_results}
\end{table}
In contrast, the FIRE prompt significantly improves performance, reducing the EER and increasing the TMR across both datasets. Notably, on the synthetic dataset, TMR@FMR=0.01 improves from 10.00\% to 60.00\%, demonstrating a substantial gain in discriminative capability. This improvement can be attributed to the step-wise reasoning process introduced by FIRE, which guides the model to focus on relevant structural features such as crease patterns, spacing, and texture.
These results confirm that prompt design plays a critical role in enabling reliable zero-shot biometric verification. Domain-aware structured prompts such as FIRE improve both prediction accuracy and interpretability.

\subsection{Prompt Sensitivity and Hallucination Analysis}
To further investigate prompt sensitivity, we evaluate a Forehead-Aware General Prompt that provides high-level instructions without structured reasoning constraints. Under this setting, GPT-4o frequently produces unstable predictions, incorrect classifications, and refusal responses.

Table 2 presents representative failure modes, including incorrect classifications, false matches, dissimilarity conflicts, and prediction instability. These errors typically arise when the model relies on global appearance cues rather than discriminative forehead-crease characteristics.

In contrast, FIRE enforces a structured feature-level analysis and substantially reduces these hallucinations. The results demonstrate that domain-aware prompting is essential for reliable zero-shot biometric verification and that prompt design directly influences both prediction accuracy and explanation quality.
\begin{tcolorbox}[
    colback=gray!4,
    colframe=blue!30!white,
    title=\textbf{Forehead-Aware General Prompt},
    coltitle=black,
    fonttitle=\bfseries,
    boxrule=0.5pt,
    arc=3mm,
    width=\textwidth,
    breakable
]
You are a visual forensics expert trained to evaluate biometric consistency. These two images are of forehead region and were likely taken under different conditions/expressions. Please provide a classification (Match or Non-Match) along with dissimilarity scores(0-100\%) and explain the results.
\end{tcolorbox}
The hallucination analysis in Table~\ref{hallulication} highlights the limitations of generic prompting. Several failure modes are observed, including incorrect classifications, false matches, instability across similar inputs, and dissimilarity conflicts.

\begin{table}[htbp]
\centering
\scriptsize
\renewcommand{\arraystretch}{1.1}
\setlength{\tabcolsep}{4pt}
\caption{Examples of Hallucinations (H) due to Incorrect Classification (IC), False Match(FM), Dissimilarity conflict(D) from GPT-4o on forehead crease-based biometric verification using Forehead-Aware General Prompt.}
\begin{tabular}{|c|c|c|c|c|}
\hline
\textbf{Image} & \textbf{GT} & \textbf{GPT } & \textbf{D} & \textbf{Hallucination } \\
\textbf{Pair} & \textbf{} & \textbf{} & \textbf{} &  \\
\hline
\multicolumn{5}{|c|}{\textbf{General Prompt Response's}} \\
\hline
P1  & Match & -- & -- & Incorrect classification  \\
\hline
P2  & Match & -- & -- & False Match, \\
& & & & Feature Hallucination\\
\hline
P3 & Match & -- & -- & Instability  \\
\hline
P4 & Match & -- & -- & Dissimilarity conflict \\
\hline
\multicolumn{5}{|c|}{\textbf{FIRE Prompt Response's}} \\
\hline
P5  & Non-Match & Match & 75\% & False Match \\
  &  &  &  & Instability \\
\hline
P6  & Match & Non-Match & 46.67\% & Dissimilarity conflict \\
\hline
P7  & Match & Non-Match & 65\% & Dissimilarity conflict \\
\hline
\end{tabular}
\label{hallulication}
\end{table}

For example, in pair P1, a genuine match is incorrectly classified due to sensitivity to lighting and pose variations. In P2, the model produces a false match by overemphasising non-discriminative visual cues. In P3, instability is observed even for visually similar images, indicating inconsistency in feature interpretation. Similar issues are observed in P4--P7, where subtle variations in illumination and crease visibility lead to incorrect decisions.

These observations demonstrate that generic prompts fail to guide the model towards consistent feature-level reasoning. In contrast, FIRE reduces such hallucinations by enforcing structured analysis, improving both stability and reliability.

\subsubsection{Stability and Reliability Analysis of FIRE}
To evaluate the robustness of FIRE, each image pair is processed through three independent stateless inference runs. Table~\ref{stability} reports the mean and standard deviation of the resulting dissimilarity scores. The observed standard deviation remains below 1.0 for both datasets, indicating limited sensitivity to stochastic LLM sampling. These findings demonstrate that FIRE produces stable and repeatable outputs while improving verification accuracy through structured reasoning.

The results show low variance across repeated runs, with consistency scores of $68.33 \pm 0.49$ for real data and $70.04 \pm 0.86$ for synthetic data. The low standard deviation indicates that FIRE produces stable and repeatable outputs despite the stochastic nature of LLM inference.
\begin{table}[htbp]
\centering
\scriptsize
\renewcommand{\arraystretch}{1.1}
\setlength{\tabcolsep}{4pt}
\caption{Reliability and Stability Analysis of FIRE Prompt Responses (Dissimilarity score)}
\label{tab:analysis}
\begin{tabular}{|c|c|c|}
\hline
\textbf{Dataset} & \textbf{Mean $\pm$ STD} & \textbf{EER\%} \\
\hline
Real & $68.33 \pm 0.49$ & 18.22 \\
\hline
Synthetic & $70.04 \pm 0.86$ & 25.44 \\
\hline
\end{tabular}
\label{stability}
\end{table}

\begin{figure}[!t]
\centering
\includegraphics[width=\textwidth]{figures/box_plot_paper.png}
\caption{Box plot showing the distribution of simulated values based on the reported mean and standard deviation for Real and Synthetic datasets. The Real data has a lower mean (68.33) and smaller variability (std: 0.49), while the Synthetic data shows a slightly higher mean (70.04) and larger variability (std: 0.86). This highlights the increased dispersion in synthetic data compared to real data.
}
\label{box_plot}
\end{figure}

Furthermore, the slightly higher variability observed in synthetic data reflects increased domain complexity, but the overall stability remains within acceptable limits. These findings confirm that structured prompting not only improves accuracy but also enhances reliability in zero-shot verification.

\subsubsection{Interpretable Verification: GPT-4o vs. Human Observers}

We further analyse the interpretability of FIRE by comparing GPT-4o outputs with human judgments on challenging image pairs, as illustrated in Figures~\ref{fig:pairs} and \ref{responses_fig}.
In many cases, GPT-4o aligns closely with ground truth and human interpretation, particularly when crease patterns are clearly visible and consistent. The model effectively identifies features such as wrinkle continuity, spacing, and symmetry, producing explanations that are intuitive and interpretable.
\begin{figure}[htbp]
\centering
\includegraphics[width=\textwidth]{figures/image_pairs_P1_P7.png}
\caption{Illustration of selected image pairs (P1 to P7) used in the qualitative analysis. These pairs highlight various failure cases, including false matches and false non-matches, under different visual conditions such as lighting changes, pose variation, texture inconsistencies, and reflection artifacts.}
\label{fig:pairs}
\end{figure}

\begin{figure*}[htbp]
\centering
\includegraphics[width=\textwidth, height=20cm]{figures/Diagram-response_supplementry.jpg}
\caption{Qualitative analysis of FIRE responses on representative verification pairs. The examples compare model predictions, confidence estimates, dissimilarity scores, and generated explanations with human interpretation, illustrating both successful reasoning cases and common failure modes.}
\label{responses_fig}
\end{figure*}
However, in challenging scenarios involving subtle variations in lighting, expression, or crease depth, the model occasionally produces incorrect classifications. These errors often arise from over-reliance on global shape similarity or insufficient sensitivity to fine-grained texture differences.

We qualitatively compare FIRE explanations with human interpretation on challenging verification pairs. GPT-4o often identifies discriminative forehead-crease characteristics such as topology, spacing, and continuity, producing explanations that align with human reasoning. However, under challenging imaging conditions, the model occasionally relies on global appearance cues and produces incorrect decisions, highlighting both the interpretability benefits and limitations of MLLM-based biometric verification.